\documentclass[journal]{IEEEtran}

\usepackage[numbers]{natbib}

\usepackage{lineno}
\usepackage{hyperref}
\usepackage{empheq}
\usepackage{url}
\usepackage{amsmath}
\usepackage{graphicx}
\usepackage{subcaption}
\usepackage{booktabs}

\usepackage{amsmath,amssymb}
\def\Var{{\rm Var}\,}
\def\E{{\rm E}\,}
\def\Std{{\rm Std}\,}
\begin{document}
\title{Estimating the volume of the left ventricle from MRI images using deep neural networks}

\author{Fangzhou~Liao,
        Xi~Chen,
        Xiaolin~Hu,~\IEEEmembership{Senior Member,~IEEE}
        and~Sen~Song% <-this % stops a space
\thanks{Fangzhou Liao and Sen Song are with the School of Medicine, Tsinghua University, Beijing 100084, China.}

\thanks{Xi Chen and Xiaolin Hu are with the State Key Laboratory of Intelligent Technology and Systems, Tsinghua
	National Laboratory for Information Science and Technology (TNList),
	and Department of Computer Science and Technology, Tsinghua
	University, Beijing 100084, China. (email: xlhu@tsinghua.edu.cn).
}
\thanks{This work was supported in part by the National Basic Research Program (973 Program) of China under Grant 2013CB329403, and in part by the National Natural Science Foundation of China under Grant 61273023, Grant 91420201, and Grant 61332007.}
}


\markboth{Journal of \LaTeX\ Class Files,~Vol.~14, No.~8, August~2015}%
{Shell \MakeLowercase{\textit{et al.}}: Bare Demo of IEEEtran.cls for IEEE Journals}



\maketitle

\begin{abstract}
    Segmenting human left ventricle (LV) in magnetic resonance imaging (MRI) images and calculating its volume are important for diagnosing cardiac diseases. In 2016, Kaggle organized a competition to estimate the volume of LV from MRI images. The dataset consisted of a large number of cases, but only provided systole and diastole volumes as labels. We designed a system based on neural networks to solve this problem. It began with a detector combined with a neural network classifier for detecting regions of interest (ROIs) containing LV chambers. Then a deep neural network named hypercolumns fully convolutional network was used to segment LV in ROIs. The 2D segmentation results were integrated across different images to estimate the volume. With ground-truth volume labels, this model was trained end-to-end. To improve the result, an additional dataset with only segmentation label was used. The model was trained alternately on these two datasets with different types of teaching signals. We also proposed a variance estimation method for the final prediction. Our algorithm ranked the 4th on the test set in this competition.
\end{abstract}


% \fi
\IEEEpeerreviewmaketitle

\section{Introduction}
\IEEEPARstart{I}{n} cardiovascular physiology, the volume of the left ventricle (LV) is important in heart disease diagnosis. The difference of end diastole volume (EDV) and end systole volume (ESV), that is EDV-ESV, reflects the amount of blood pumped in one heart beat cycle (Stroke Volume, SV), which is often used in the diagnosis of heart diseases. Another frequently used indicator is the ratio (EDV-ESV)/EDV (Ejection Fraction, EF).

Modern medical imaging methods such as ultrasound, radiology and MRI make it possible to estimate the volume of LV. The first step is to acquire a set of slices along the $z$-axis (the long axis of the LV). Then experienced doctors draw contours on the slices to get the area of LV in each slice and calculate volume by accumulating the areas along the $z$-axis. This step is time-consuming. Usually, an MRI stack costs a senior doctor more than ten minutes in the analysis. If an automatic method is available, it will speed up the diagnosis process.

%The effort of designing such algorithms started two decades ago \citep{devereux_echocardiographic_1977,dodge_usefulness_1966,fortuin_determination_1971} when only radiographic and echocardiographic methods were available. During the 1980s a new technique, MRI, developed quickly. It has high spatial resolution and can obtain slice stacks quickly, which makes it possible to reconstruct the dynamic 3D structure of the beating heart.

A straightforward approach for solving this problem is to first segment LV area in each MRI slice, then estimate the volume of LV by integrating the areas across difference slices. Over years many LV segmentation algorithms have been proposed (e.g., \citep{lorenzo-valdes_atlas-based_2002,kaus_automated_2004,lynch_segmentation_2008}). But the progress towards solving the problem was hindered by the lack of benchmark dataset. The first large public available dataset called the Sunnybrook dataset was released in 2009 \citep{radau_evaluation_2009}, which contains hundreds of cases. It provides not only the diagnosis label for each case but also LV contours labeled by experts for some cases. It is the first systematically labeled dataset, which facilitates the application of machine learning algorithms on this problem, including deep neural network \citep{ngo_left_2013,avendi_combined_2016} and random forest \citep{dreijer_left_2013}. Traditional learning-free methods can also benefit from it \citep{hu_hybrid_2013,liu_automatic_2012}.

In 2015 Kaggle.com organized a competition named ``Second National Data Science Bowl". A dataset with hundreds of 3D MRI videos were provided with EDV and ESV labels. The task was to predict EDV and ESV in new videos \citep{kaggle_data_2015}. External data except the Sunnybrook dataset was not allowed to be used. Clearly, a single scalar volume label is quite uninformative for training an accurate model because the volume depends on the classification of every pixel in each slice. To make the model capable of doing segmentation, we proposed to train a deep neural network using both the volume labels on the training set and the segmentation labels on the Sunnybrook dataset. The proposed pipeline was based on deep neural networks. The competition had two rounds. In the initial round there were more than 700 teams, and in the final round there were 192 teams. Our algorithm ranked the 4th in the final round. In the paper, we describe our algorithm in detail to make a contribution to related clinical practices\footnote{The source codes are released on https://github.com/lfz/Heart-Volume-Estimation.git.}.

\section{Related work}
\subsection{Deep learning models for general image segmentation}

Before the rise of deep learning, most image processing algorithms rely on sophisticated handcrafted features (e.g. HOG \citep{dalal2005histograms}, Haar \citep{viola2001rapid}, SIFT \citep{lowe_object_1999}, LBP \citep{ahonen2006face}) and complex classifiers (e.g. random forest, support vector machine, cascade classifier). Deep learning provides an end-to-end solution to these tasks because in these models both features and classifiers are learned together. Now deep learning is enjoying fast development for computer vision tasks including image classification, object detection, image segmentation and so on. See reference \cite{lecun_deep_2015} for a recent review. As image segmentation is one of the main parts in our pipeline, we briefly review existing deep learning models for general image segmentation below.

According to whether there is upsampling operations, existing deep learning models for image segmentation can be classified into two categories. The first category of models do not use upsampling because they use patches as inputs and classify their central pixels \citep{ning_toward_2005,jain_supervised_2007}. This process slides on the image to predict the label of every pixel. It is slow because every single patch requires a feedforward computation. However, recently this problem was partially solved by rarefying the weight \citep{yu_multi-scale_2016,lee_recursive_2015}.

The other category of models takes an image as input and output a label map. They usually use deconvolution or unpooling to upsample the Low-resolution, high-level representation of an image back to the original size. For example, the fully convolutional network (FCN) \citep{long_fully_2015} uses a pre-trained network as an encoder, then learns a decoder at different levels and combine them together to achieve finer result. In the hypercolumns network \citep{hariharan_hypercolumns_2015}, features in different levels are concatenated together, and segmentation is conducted based on this new layer with multi-scale information. The U-Net \citep{ronneberger_u-net_2015} introduces an U-shape network architecture which is more efficient in capturing fine-scale information. The deconvolution network \citep{noh_learning_2015} stores indexes in the max-pooling layer in the encoding step, and uses them in corresponding unpooling layers in the decoding step. An advantage of these models is their fast prediction due to the fully convolution style.

\subsection{MRI image segmentation}
MRI is currently an important imaging technique for cardiology and its data is widely used in segmentation algorithm development. \citet{lorenzo-valdes_atlas-based_2002} built a standard heart atlas and registered every slice stack to it. \citet{kaus_automated_2004} used a deformable model to build 3D model first and segmented the inner surface in 3D space. \citet{lynch_segmentation_2008} used a level-set method and took temporal information into consideration. \citet{eslami_segmentation_2013} proposed a guided random walk method. Instead of finding the LV chamber, they targeted for segmenting the ventricle wall. \citet{nambakhsh_left_2013} proposed a convex relaxed distribution matching algorithm which finds the best segmentation that matches the intensity/shape prior distribution. \citet{zhang_4-d_2010} adopted active appearance and shape models to build a general model of the heart shape.

In recent years, researchers began to apply deep learning algorithms to LV segmentation problem. \citet{ngo_combining_2017} used a deep belief network (DBN) and combined it with level-set method \citep{osher_fronts_1988}. \citet{avendi_combined_2016} combined a multilayer perceptron with deformable model \citep{kass_snakes:_1987}.
%However, network structure in both works were too simple, which limited their performance. On the other hand because
One reason for that both works incorporated traditional methods (level set method and deformable model) is that the models were trained on the Sunnybrook dataset, which is too small to train complex and powerful models. Given this, \citet{tran_fully_2016} collected multiple datasets and trained an end-to-end model (FCN) on them.

Though the task in the ``Second National Data Science Bowl'' competition organized by Kaggle.com was to estimate the volume of LV, many teams (including our team) chose to segment LV in every slice of MRI images as an important step. The first place and third place teams adopted this strategy and used extra hand-labeled data to increase the number of training samples for segmentation task \cite{tencia_summary_????,wit_3rd_????}, which we think is the primary reason for their better performance than our model. The second place team trained end-to-end models to directly predict the volume value without using segmentation labels. The final result was based on an ensemble of 44 models \cite{ira_diagnosing_????}, yet we just used a single model. All of the top four teams built their models based on deep convolutional neural network.

\section{Volume estimation}

\subsection{Data preprocessing}\label{section_Data}
\begin{figure}
	\begin{subfigure}[b]{\columnwidth}
		\includegraphics[width=\columnwidth]{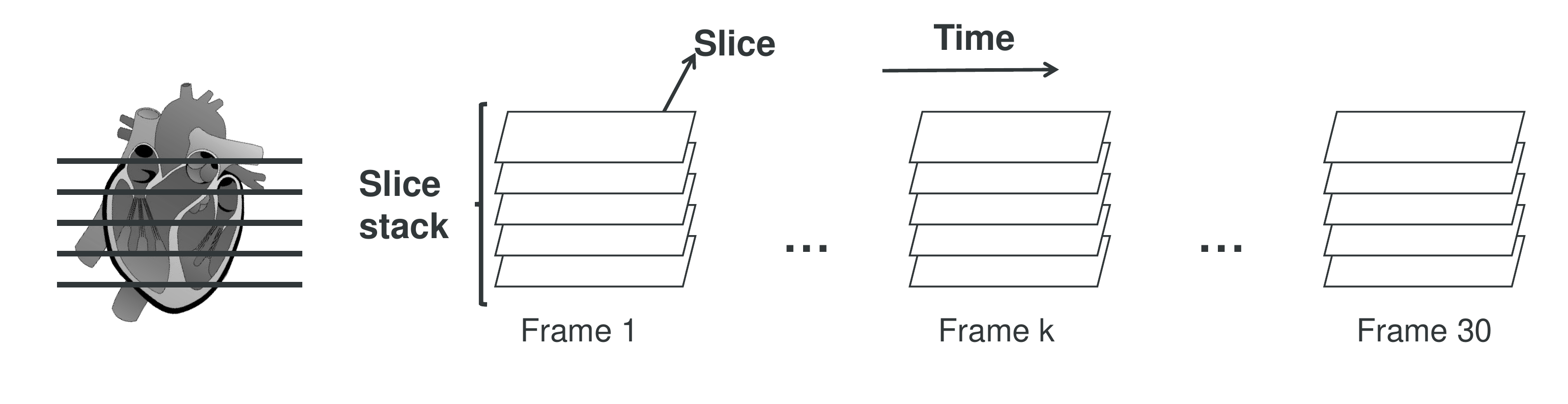}
		\caption{}
		\label{rawdata}
	\end{subfigure}
	
	\begin{subfigure}[b]{\columnwidth}
		\includegraphics[width=\columnwidth]{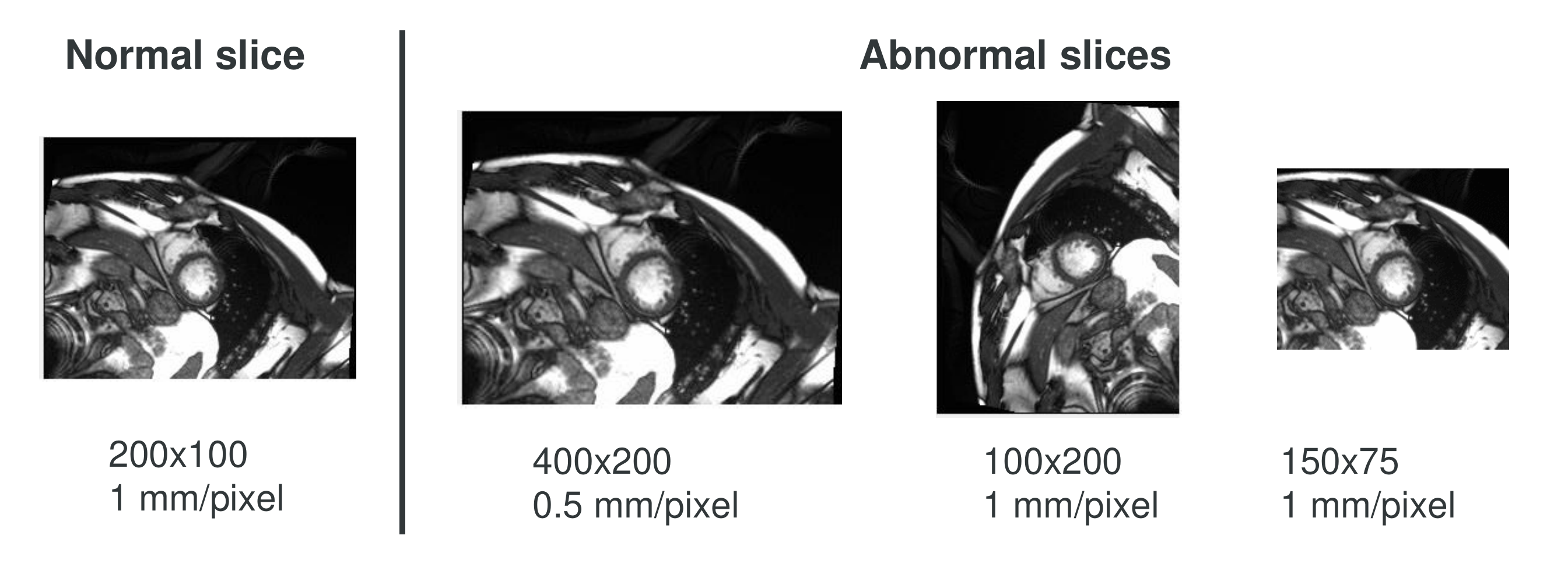}
		\caption{}
		\label{preprocess}
	\end{subfigure}
	\caption{Illustration of the raw data and pre-processing techniques. (a) Short axis views of one case. It consists of 30 frames along time, and each frame has some slices. The number of slices varies from case to case. (b) Sample slices. Most slices in one case share common properties. They are treated as normal samples. But there are other slices as shown by other three examples, which require resizing, rotation, padding/cropping, respectively, in preprocessing to make them consistent with normal slices in image size, spatial resolution, and orientation.
		%When there exist an image that has inconsistent property (spatial resolution, angle, image size), corresponding process will be conducted to convert it to property shared by most slices.
	}
\end{figure}

The training, validation, and test set consist of 500, 200, and 440 cases respectively. Each case has several views: short axis view (sax), 4-chambers view (4-ch), and 2-chambers view (2-ch). Each view is a 4D data: $x,y$-axis (2D frame) $\times$ $z$-axis (slice location) $\times$ $t$-axis (imaging time). We only used the sax view data in this competition. Each slice location is usually sampled 30 times, which correspond to 30 frames, in a heart-beating cycle (\autoref{rawdata}). In this paper a \textit{frame} may refer to a slice stack (3D) or a slice (2D) according to the context. To get better results, we used 630 cases for training and 70 cases for validation.

On the training set and validation set, two volume values of LV were provided for each case corresponding to ESV and EDV respectively, but their corresponded end of systole frame (ESF) and end of diastole frame (EDF) were not provided. We identified the two frames for each case with visual inspection which had minimum and maximum volumes, respectively. Then each case had two frames with volume labels. They were used in the volume fine-tuning part (\autoref{sec_vol}).

In some cases, spatial resolution, orientation, and size can be different from slice to slice (\autoref{preprocess}). That information is stored in metadata of each dicom file. According to these information, the inconsistent slices were rotated/resized/cropped/padded to be consistent with most slices.

We used Sunnybrook dataset as a supplementary dataset. It also consists of three sets: training, validation, and test. It has human-labeled LV contour (both inner contour and outer contour) for some cases which were used in this competition. We combined these three sets together and extracted 400 slices with outer contours of LV for the detection task (\autoref{sec_roiproposal}). We also extracted 725 slices with inner contours of LV for the segmentation task (\autoref{sec_segment}).

\subsection{Overview of our method}

\begin{figure}
	\centering
	\includegraphics[width=\columnwidth]{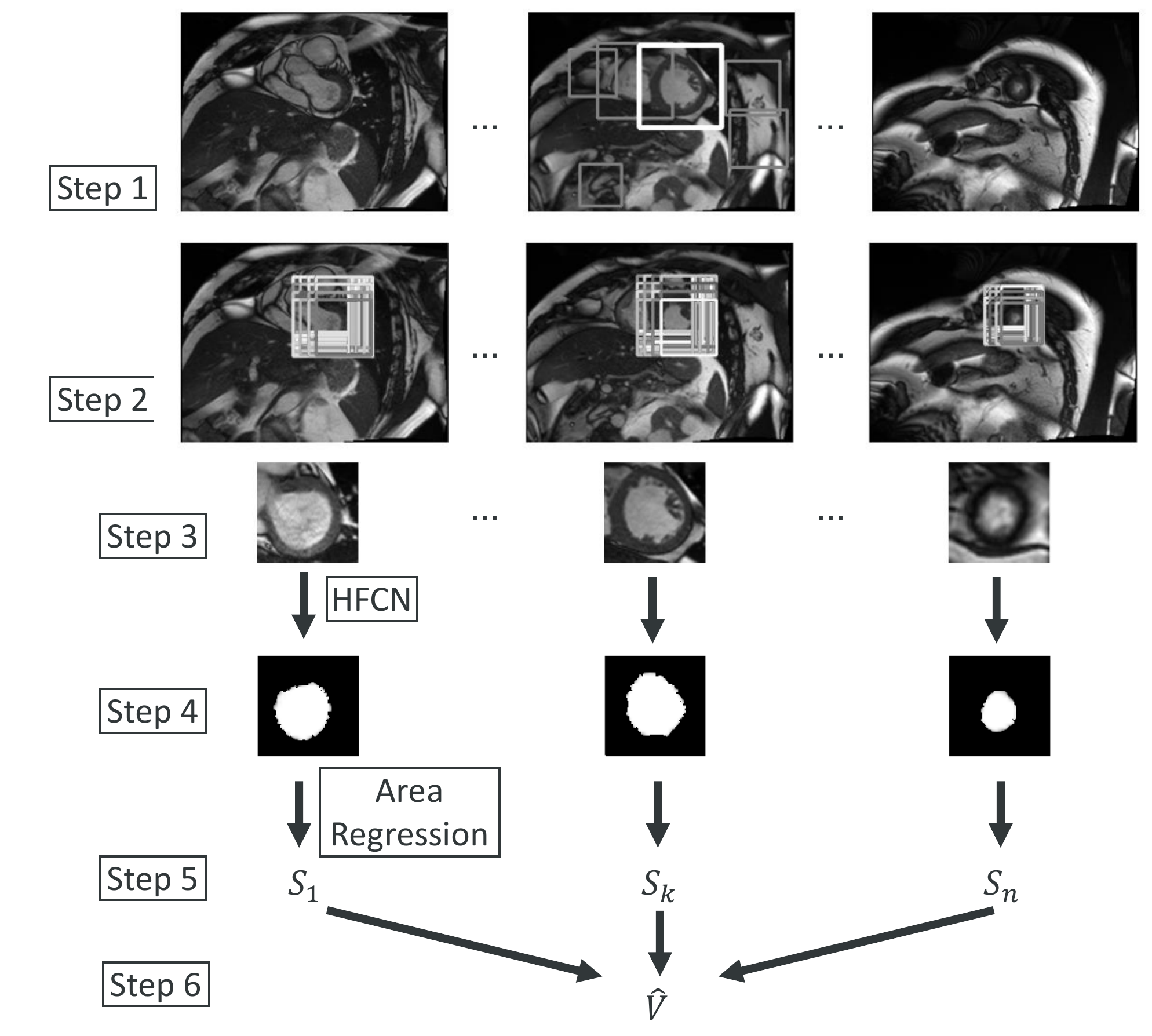}
	\caption{The pipeline of LV volume prediction for one frame. Step 1: LBP detector gives some ROI proposals in the middle slice, and a classifier is used to determine the correct one. Steps 2 and 3: The ROI in the middle slice might be too large for other slices, so we choose the best one among many shifted and resized ROIs (the gray squares). Step 4: Each ROI is fed to a network (HFCN) for segmentation. Step 5: Calculate area of LV in each slice based on the segmentation results. Step 6: Estimate volume based on the LV area in each slice. }
	\label{pipeline}
	
\end{figure}

We first show the overview of the proposed pipeline (\autoref{pipeline}), then in subsequent sections we describe the individual steps in detail.

First, we trained an LBP \citep{liao_learning_2007} cascade detector to produce some ROI proposals about the location of LV in the middle slice of the first frame. Since the visual appearance of LV was prominent (a concentric circle), especially in the middle slice, the correct ROI was always included in the proposals. Then a CNN was applied to classify those proposals and the one with the highest score was chosen as the ROI. See \autoref{pipeline}, Step 1.

Second, we extended the ROI to all frames and all slices, and segmented the LV in each ROI and calculated the area based on a modified FCN, called hypercolumns fully convolutional network (HFCN, \citep{hariharan_hypercolumns_2015}). This model was trained on the Sunnybrook dataset. See \autoref{pipeline}, Steps 2 to 4.

Finally, for each slice, we calculated areas based on the model in the previous step and got the volume by integrating areas along the $z$-axis. Since the min and max volume labels for every case (ESV and EDV, respectively) were available, the error signal could be used to fine-tune HFCN. See \autoref{pipeline}, Steps 5 to 6.

\subsection{ROI proposal in the middle slice of the first frame}
\label{sec_roiproposal}
Since LV is relatively small in each slice, performing segmentation on the whole image is very challenging. It would be nice if we could have a finer ROI bounding the LV in every slice. Since LV, like a spindle, is big at the middle and small at the two ends, if we can localize the ROI in the middle slice, then this bounding box should also bound LV in other slices.

Face detection is a similar problem and has been studied for years in computer vision. We chose a very simple algorithm used in face detection: built a cascade classifier based on LBP features. We extracted 400 positive ROIs from the Sunnybrook dataset and semi-automatically labeled 500 positive ROIs in the training dataset, then trained the cascade classifier with ten times more negative samples.

Since concentric circle may appear in other locations in the body, this approach produced some false-positive ROIs. But the correct ROI was among the proposals in almost all cases. And this bounding box was usually tight enough. Another classifier was then used to choose the correct ROI. See below.
\begin{figure}
	\centering
	\begin{subfigure}[t]{.4\columnwidth}
		\centering
		\includegraphics[width=.8\columnwidth]{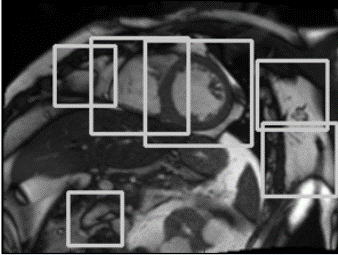}
		\caption{}
		\label{lbp}
	\end{subfigure}%
	\begin{subfigure}[t]{.4\columnwidth}
		\centering
		\includegraphics[width=.8\columnwidth]{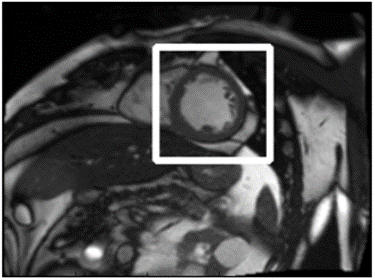}
		\caption{}
		\label{classifyresult}
	\end{subfigure}
	
	\centering
	\begin{subfigure}[t]{.4\columnwidth}
		\centering
		\includegraphics[width=.8\columnwidth]{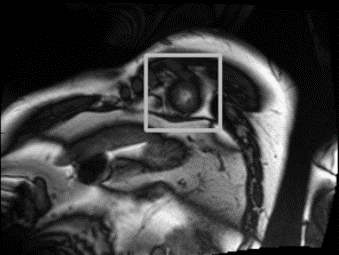}
		\caption{}
		\label{beforerefine}
	\end{subfigure}%
	\begin{subfigure}[t]{.4\columnwidth}
		\centering
		\includegraphics[width=.8\columnwidth]{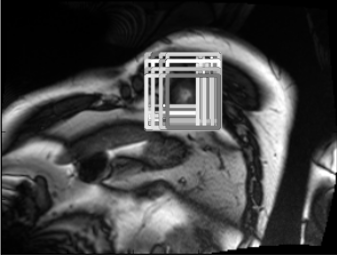}
		\caption{}
		\label{refine}
	\end{subfigure}%
	\begin{subfigure}[t]{.2\columnwidth}
		\centering
		\includegraphics[width=.8\columnwidth]{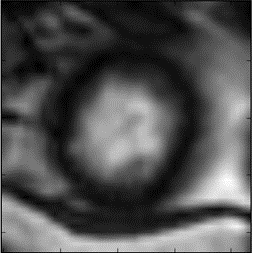}
		\caption{}
		\label{refineresult}
	\end{subfigure}%
	
	\caption{ROI extraction in the first frame. (a, b) Extraction in the middle slice of the first frame. First, LBP detector finds many ROI proposals in the middle slice (a). Then all proposals are fed to a CNN classifier which outputs a confidence score for every proposal. The proposal with the highest score is selected (b). (c-e) ROI refining process in other slices. Assuming that we have determined the bounding box of one slice and move to the next. First, this slice inherits the bounding box from the last slice as the initial guess (c). Many ROI proposals with different sizes and positions (the gray boxes) are extracted and tested with the classifier (d), and the one with the highest confidence score is chosen (e).}
\end{figure}

\subsection{ROI classification in the middle slice of the first frame}
\label{sec_roi_classify}
The most distinctive feature of the heart lies in the spatial-time domain: the heart is beating. Yet it can not be captured by the method described above. To utilize the information of ``heart beating", for each case, 6 frames were evenly chosen from 30 frames in the time series and concatenated together, so that the top and bottom channel images were diastole and the middle channel image was systole. Then all proposals given by the LBP detector were extended to these frames, resized to $25 \times 25 \times 6$ and manually labeled as correct or wrong. Then a 3-layers CNN was trained to perform classification (\autoref{roiclassifier}). As a simple way of data augmentation, all positive samples were rotated by $ 90^\circ, 180^\circ, 270^\circ$.
After trained in 500 cases, the detection system obtained 100\% accuracy on the validation set (\autoref{lbp},\ref{classifyresult}).

% \usepackage{booktabs}
% \usepackage{graphicx}
\begin{table*}[]
	\centering
	\caption{The configuration of CNN for ROI classification.}
	\label{roiclassifier}
	\resizebox{\textwidth}{!}{%
		\begin{tabular}{@{}ccccccccc@{}}
			\toprule
			Layer & Data & Conv1 & Pool1 & Conv2 & Pool2 & Conv3 & Pool3 & Classifier \\ \midrule
			Param & \begin{tabular}[c]{@{}c@{}}Mean\\   abstraction\end{tabular} & \begin{tabular}[c]{@{}c@{}}5x5,\\   Pad 2\\   St 1, BN, ReLU\end{tabular} & \begin{tabular}[c]{@{}c@{}}Max, 2x2\\   St 2, Drop 0.4\end{tabular} & \begin{tabular}[c]{@{}c@{}}5x5, Pad 2\\   St 1, BN, ReLU\end{tabular} & \begin{tabular}[c]{@{}c@{}}Max, 2x2\\   St 2, Drop 0.4\end{tabular} & \begin{tabular}[c]{@{}c@{}}3x3,\\   Pad 1\\   St 1, BN, ReLU\end{tabular} & Ave, 6x6 & Softmax \\
			Size & 25x25x6 & 25x25x32 & 12x12x32 & 12x12x32 & 6x6x32 & 6x6x32 & 1x1x32 & 1x1x2 \\ \midrule
			\multicolumn{9}{l}{St: stride, BN: batch normalization, Drop: dropout ratio, Ave/Max: average/max pooling}
		\end{tabular}%
	}
\end{table*}

\subsection{ROI localization in other slices and other frames}
The previous step gives an accurate bounding box at the middle slice of the first frame. The simplest way to have ROI in all frames and slices is to copy the result from this slice. It is acceptable in the time-domain as LV does not change too much in the heart-beating cycle. But because the shape of LV is like an upright spindle, the bounding box in the middle slice is too large for other slices. So a refinement step for the bounding box of other slices in the first frame is necessary.
%Begin with the slice above the middle one, each slice inherents ROI in the below slice as initial guess, and get its own based on that, and this procedure is repeated in the slices below the middle one.
Starting from the middle slice, we extended the ROI to other slices sequentially in bottom-up and top-down directions respectively. Each slice inherited the ROI of its neighboring slice determined in the last step as the initial guess and refined it with the method described below.
Many square patches were extracted around the initial guess with different corner point whose length ranges from 0.6 to 1.1 times the length of the initial guess (similar to sliding window, \autoref{refine}). For each patch, we chose 6 frames in time series and concatenated them as described in \autoref{sec_roi_classify}, then fed them to the 3-layer CNN mentioned above (\autoref{roiclassifier}), which gave us a confidence score for every patch. The patch with the highest score was the refined ROI. In some cases where the slice location was beyond the range of LV in the $z$-axis so that LV was absent or the shape of LV is abnormal, the classifier would give a low score for every patch. Under such circumstances, we chose the ROI in its nearest slice as its own.

After the bounding boxes in all slices in the first frame were obtained, they were extended to corresponding slices in other frames directly.

\subsection{Segmentation}
\label{sec_segment}
After refining ROIs, the target area usually occupied 20-50\% of the ROI. All ROIs were resized to $48 \times 48$. The next step was to segment LVs and estimate their areas in individual slices.

\begin{figure*}
	\centering
		\centering
		\includegraphics[width=0.7\textwidth]{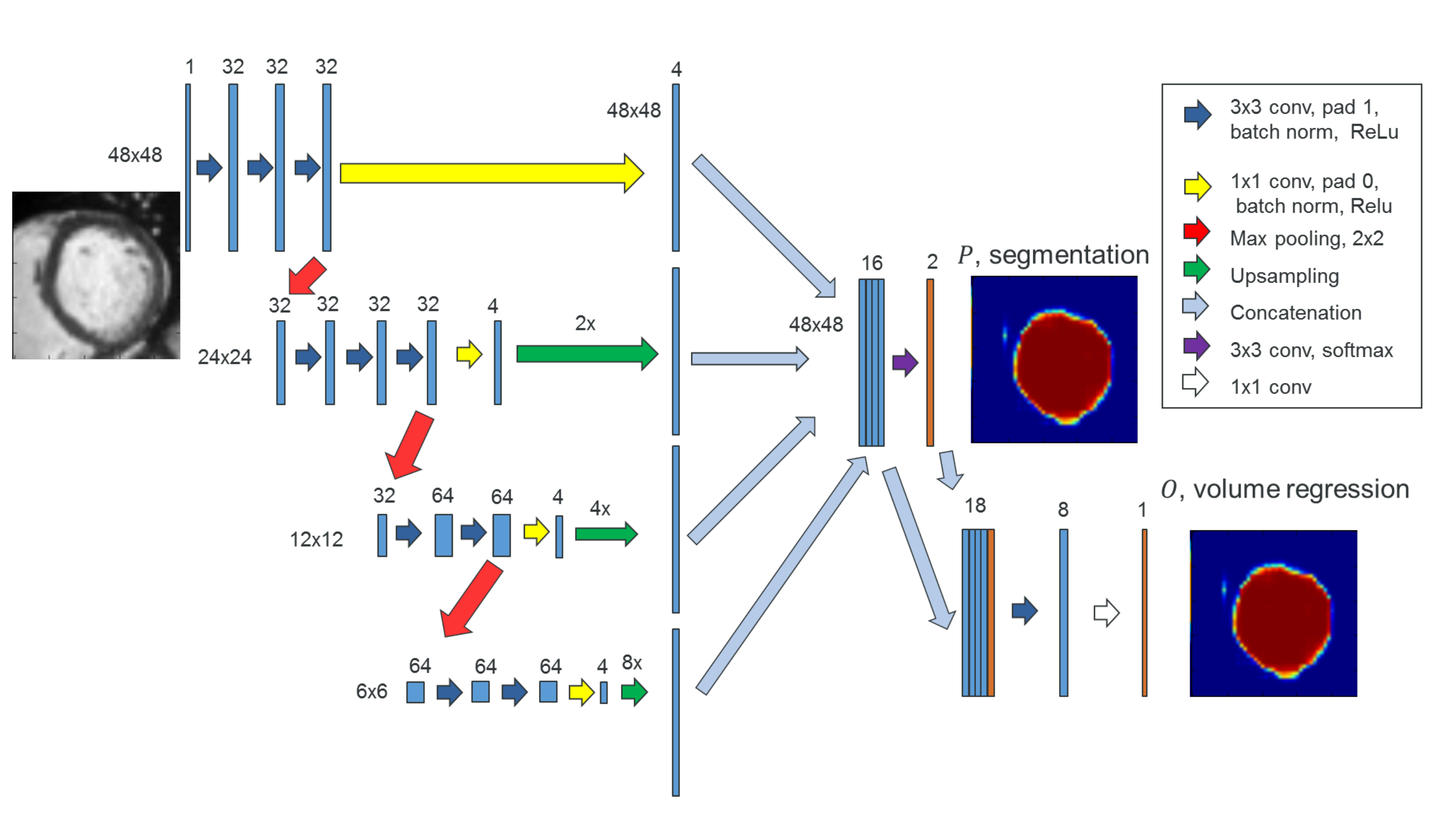}
	\caption{The architecture of HFCN. Best viewed in color.}
		\label{network}
\end{figure*}
We first tried the fully convolutional network (FCN) \citep{long_fully_2015} for segmentation. It received a 2D image as input and gave a binary segmentation result. The training set was obtained from the Sunnybrook dataset. Although FCN uses multiscale information, its predictions in different levels are independent. The final result is just an average of those predictions. It is an inefficient way of integrating multiscale information. Our experiments validated this point. To overcome this shortage, we adopt the idea of hypercolumns \citep{hariharan_hypercolumns_2015}. Features from different levels were concatenated to form a new layer, and segmentation was based on this new layer (\autoref{network}). This model is termed hypercolumns FCN (HFCN) in this paper.

\subsection{Volume estimation}
\subsubsection{Naive volume calculation}
After training HFCN, we fed the ROI in every slice to it and got an output image whose pixel value was the likelihood of being part of LV. By binarizing and averaging all pixel value, we could get the area fraction of LV in this slice. As the pixel resolution and image size was known, the physical area of LV could be deduced. The whole process is formulated as follows:

\begin{subequations}\label{grp}
	\begin{empheq}[left=\empheqlbrace]{align}
	P&= f_w(I)\label{prob},\\
	B&= \text{binarize(P)}\label{bianrize},\\
	F&=\text{mean}(B)\label{area fraction},\\
	S&=F \cdot (rL)^2\label{physical area},
	\end{empheq}
\end{subequations}
where $I$ is the input ROI, $P$ is the output image whose pixel value is the probability of being part of LV, $w$ is parameters of HFCN, $F$ is area fraction, $r$ is physical length of a pixel, $L$ is patch length, and $S$ is physical area. These steps can be combined in one step:

\begin{equation}
S = g_w(I).
\label{singleS}
\end{equation}

The next step was to calculate the volume based on the areas at different locations in the $z$-axis. We assume the cross section of LV is always a circle. Then the volume between two neighboring slices is a truncated circular cone, whose volume is
\begin{equation}
V_i = (S_i + S_{i+1} + \sqrt{S_i S_{i+1}})(L_{i+1}-L_{i})/3,
\label{singleV}
\end{equation}
where $S_i$ is the area of LV in the $i$-th slice and $L_i$ is the coordinate of the $i$-th slice in the $z$-axis. The total volume of LV is the sum of all truncated cones:
\begin{equation}
\hat{V} = \sum_{i=1}^{N-1}V_i,
\label{totalV}
\end{equation}
where $N$ is the number of slices.

\subsubsection{Fine-tuning with volume labels}
\label{sec_vol}
If the segmentation results were perfect, the volume given by Equation \eqref{totalV} was exact. Unfortunately, the segmentation network did not work perfectly partly due to the small size of the training set. To augment the dataset, every case was augmented by rotating $ 90^\circ, 180^\circ, 270^\circ$. To further improve the results we needed to utilize the volume labels of the dataset provided in this competition to fine-tune the segmentation network.

We chose $l1$-norm loss function instead of the usually used $l2$-norm loss function:
\begin{equation}
L = |V-\hat{V}|,
\label{eq_loss}
\end{equation}
where $V$ is the ground truth volume.
The reason is as follows. The labels were variant due to subjective standards of different doctors, even misleading in some cases\footnote{For instance, the sax view data of the case 429 in the training set, and cases 595 and 599 in the validation set were incomplete, though the 4-ch view data of them were complete. The labels of these cases, though might be correct as doctors could estimate the volume in the 4-ch view, were misleading to our algorithm, since it worked solely on the sax view. }. It is well-known that the $l1$-norm loss is more robust to outliers than $l2$-norm loss. Another reason was that the $l1$-norm loss is more similar to the evaluation function used in the competition, which will be discussed in \autoref{evaluation part}.

To this end, the estimated $\hat{V}$ should be a differentiable function of the network parameters $w$. Note that all Equations \eqref{grp} to \eqref{totalV} meet this requirement except Equation \eqref{bianrize}. One solution is substituting it with a sigmoid activation function: $B = \text{sigmoid}(P).$ But because the output of the sigmoid function is always between 0 and 1, a consistent positive bias for non-LV pixels and negative bias for LV pixels will be caused. In our algorithm. We used another method instead.

We concatenated $P$ with its previous layer to form a new layer. Then a $1 \times 1$ convolutional layer was added after it (see \autoref{network}). This was essentially a linear regression. Surprisingly, we found that it worked better than the sigmoid transformation.

Denote the new output by $O$, and the new set of parameters by $\hat{w}$:

\begin{equation}
O=\hat{f}_{\hat{w}}(I).
\label{newO}
\end{equation}
Then Equation \eqref{area fraction} becomes
%The new output $O$ is no longer biased, so now we can get LV fraction by directly averaging it:

\begin{equation}
F = \text{mean}(O).
\end{equation}

%However, the result does not turn out well, because the size of training dataset in segmentation part is too small. Although we can augment it infinitely, the true information hidden in the label -- the knowledge of doctors did not improve. Besides, all data are positive sample. In other words, it never learns how to distinguish those ambiguous false positive image. Thus the final result is always bigger than label.

%To overcome this shortage, we need to incorporate the volume label information in our algorithm to fine-tune the parameters. From \autoref{singleV}\eqref{totalV}, $V$ could be written as differentiable function of $S_i$, which is function of $w$. So if $S_i$ is differentiable to $w$, the final $V$ is also differentiable to $w$. Then could we define a loss function and minimize it by optimizing $w$. But because the existence of \autoref{bianrize}, the assumption is wrong. We will discuss how to solve it in next section.

%Another modification is instead of using \autoref{singleV}, we adopted a less precise but more robust volume estimation method:

Though Equation \eqref{singleV} is differentiable, the presence of $\sqrt{S_i S_j}$ makes training unstable. We empirically found that substituting it with the arithmetic mean $\frac{S_i + S_j}{2}$ made learning more robust. In other words, Equation \eqref{singleV} is changed to
\begin{equation}
V_i = (S_i+S_{i+1})(L_{i+1}-L_i)/2.
\label{singleV2}
\end{equation}

\begin{figure*}
	\centering
	\includegraphics[width=0.7\linewidth]{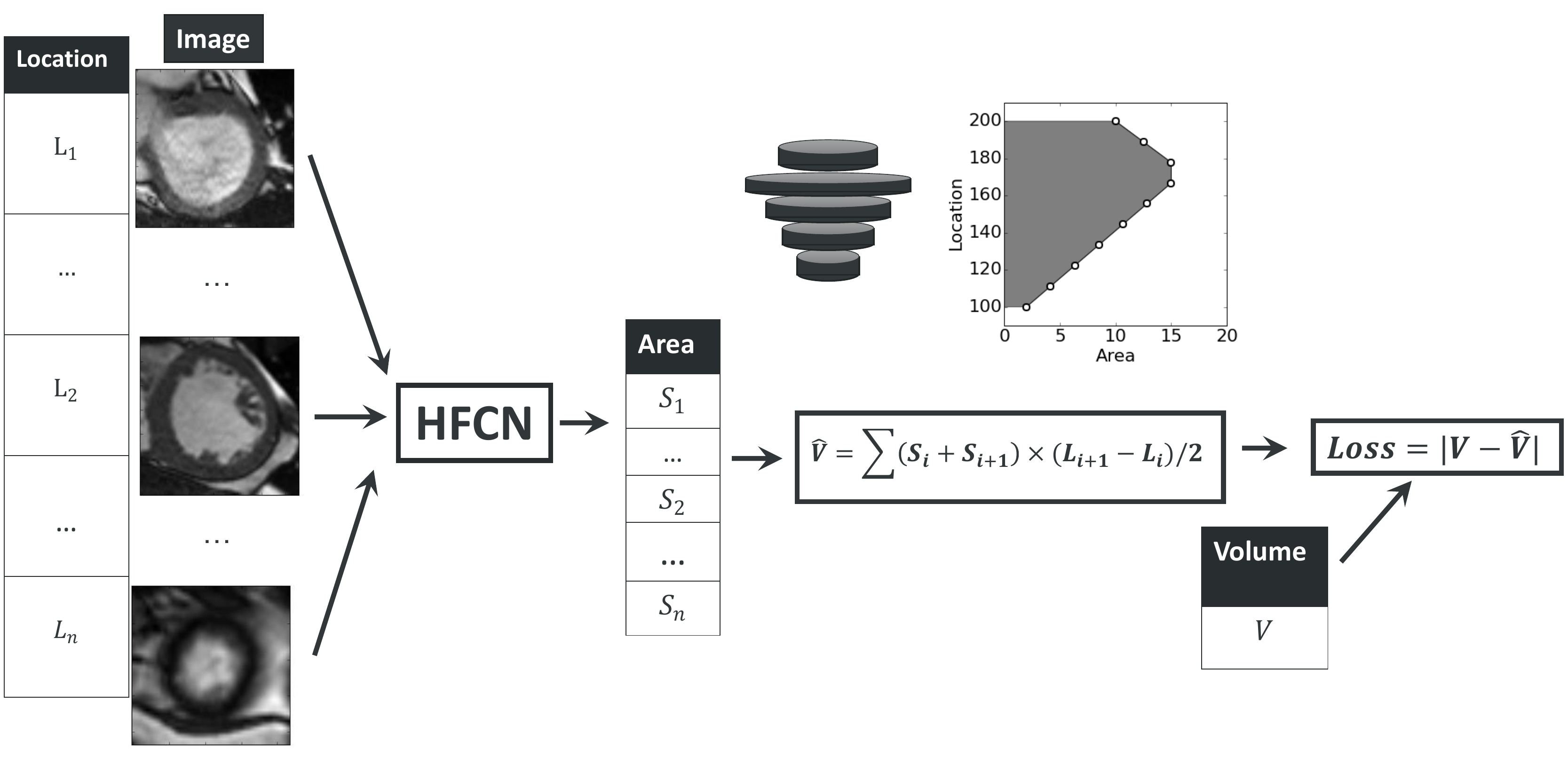}
	
	\caption{The diagram of volume fine-tuning. Because the whole pipeline is differentiable, the error can be backpropaged to HFCN.}
	\label{diagram}
\end{figure*}

The whole diagram for volume estimation is shown in \autoref{diagram}.

\subsection{Alternate training}
Though our goal was to estimate the volume of LV, simply training the model using the volume labels provided by the competition organizer did not work well. This was because the estimation quality heavily relies on the image segmentation results, but the organizer did not provide ground-truth labels for segmentation. Our solution was to alternately train the model on the competition dataset with volume labels and on the Sunnybrook dataset with image segmentation labels.
%To sum up, there are two separate tasks in our algorithm. The first one is a image segmentation task, in which data and label comes from Sunnybrook dataset. The second one is volume regression task, in which data and label comes from the training data provided in competition. But these two tasks share a segmentation model.

In other words, both two outputs of HFCN network (see \autoref{network}) were used in training. The first one, $P$, was used in the segmentation task, and the second one, $O$, was used in the volume regression task. These two tasks were trained alternately. Specifically, a training block was composed of a segmentation epoch and a volume regression epoch. This scheme made the most use of limited label resources and made these two tasks regulizers of each other to reduce over-fitting. To achieve higher precision in volume regression, we removed segmentation epochs in training after 100 blocks. Then the training continued with only regression epochs until convergence. A total of 200 blocks were used in training. The learning rate was set to 0.001 with momentum of 0.9, and discounted by 0.1 every 70 blocks.
\subsection{Re-picking of ESF and EDF}
As mentioned before, we picked the ESF and EDF in the training set by visual inspection. But it was potentially inaccurate. So after training, we fed every slice of a case in the training set to the model and identified its ESF and EDF by finding the min and max volumes from model outputs. Then we re-trained the segmentation network based on the new ESF and EDF.

\section{Distribution estimation}
\subsection{Evaluation function}
\label{evaluation part}
The evaluation function in this competition was Continuous Ranked Probability Score (CRPS). Participants were required to upload a distribution of predicted volume (EDV or ESV) over all possible volumes $n\in\{0,1,2,...,599\}$ (The upper limit was set to 599 because no ground truth value was larger than this value). Say for the $m$-th case, the ground truth value is $V_m \in\{0,1,2,...,599\}$. Then the ground truth distribution is a delta function peaked at $V_m$. Clearly its cumulative distribution function (CDF) is
%the predicted cumulative distribution function (CDF) is $P_m(n)$. Naturally, the optimal result is denoted as
\begin{displaymath}
H_m(n)=\begin{cases}
0 & \mbox{if $n\le V_m$}, \\
1 & \mbox{if $n\geq V_m$}.
\end{cases}
\end{displaymath}

Denote the predicted probability density function (PDF) and its CDF for the $m$-th case by $f_m(n)$ and $P_m(n)$, respectively. The CRPS for each estimation is defined as a distance between $P_m(n)$ and $H_m(n)$:
\begin{equation}
C_{m,k}= \frac{1}{600}\sum_{n=0}^{599}(P_{m,k}(n)-H_{m,k}(n))^2,
\label{CRPS}
\end{equation}
where $k$ can be EDV or ESV. The evaluation metric is the mean CRPS over both $k$ and all $m$.

\subsection{Generation of a distribution}
Note that the proposed algorithm can only give a single value $\hat{V}$ for any case but does not give a distribution over all $n$. In the simplest case, $f(n)$ is a delta function peaked at $\hat{V}$, then $P(n)$ is a step function, and CRPS becomes $C = \frac{1}{600}|V-\hat{V}|$. This is another reason why we used $l1$-norm loss for volume fine-tuning. However, we found that the delta function is not a good choice.

To better understand the properties of this evaluation function, by assuming the distribution $f(n)$ to be the PDF of $N(V+d,\sigma^2)$, we numerically calculated the distribution of CRPS (\autoref{loss}) on $d$ and $\sigma$. We found that $\sigma$ could significantly influence performance. For example, $C(d=10, \sigma=0)=0.0167$, $C(d=10, \sigma=12)=0.0094$, which led to more than 40\% reduction. We found that for every $d$ there existed an optimal $\sigma^\star$ that minimized CRPS, and $\sigma^\star$ was approximately a linear function of $d$. But since $d$ was unknown during test, this $\sigma^\star$ was only used as a performance reference on the training set.

Let $f(n)$ be the discrete PDF of $N(\hat{V},\sigma^2)$. Since $\hat{V}$ is known the problem reduces to find an optimal $\sigma$. A natural choice of $\sigma$ is $\Std(\hat{V})$, and by numerical experiment, we found that this choice is good enough though not optimal (see Appendix). By combining Equations \eqref{area fraction}\eqref{totalV}\eqref{singleV2}, we know that $\hat{V}$ can be written as a linear summation of $F_i$:
\begin{displaymath}
	\hat{V}=\sum_{i=1}^{N}\alpha_i F_i.
\end{displaymath}
We assume that $F_i$ and $F_j$ are independent when $i\neq j$, then the variance of $\hat{V}$ is also a linear summation of variance of $F_i$:
\begin{displaymath}
\Var(\hat{V})=\sum_{i=1}^{N}\alpha_i^2 \Var(F_i).
\end{displaymath}
 $\Var(F_i)$ can be estimated from data in \autoref{reg1}. Specifically, we collected all dots with predicted area in the range $[F_i-0.05, F_i+0.05]$, and calculated the variance of their true areas. Because this variance stands for the probable dispersion of true area given the predicted area, it is used as $\Var(F_i)$.

Finally we used a modified form $\hat{\sigma}=\sqrt{\beta{\Var(\hat{V})}}$ where $\beta$ was a hyperparameter hand-tuned on the validation set. The reasons are as follows.
First, the neighboring slices are similar, so $F_i$ and $F_j$ are not independent. Second, the distribution of samples used to compute $\Var(F_i)$ (\autoref{reg1}, extracted and augmented from Sunnybrook dataset) might deviate from that of the competition dataset, so the computed variance was not the desired value.
Based on the evaluation performance on the validation set, the best $\beta$ was 0.5.

\section{Results}
\subsection{Image segmentation}
For image segmentation, we trained the model on all data of the Sunnybrook dataset including the training set, validation set, and test set. The training accuracy was 98.3\% without fine-tuning using the volume information of the Kaggle competition dataset, and this accuracy dropped to 97.2\% after alternate training on both datasets.
%so we can only report the accuracy in training stage. If only segmentation task was conducted, the accuracy is 98.3\%. And in mixed training, the accuracy drops to 97.2\% due to the reduction of over-fitting.

\begin{figure}
	\centering
	\includegraphics[width=\linewidth]{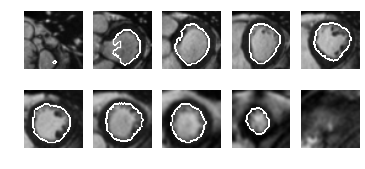}
	
	\caption{Segmentation results of all slices at a frame of one case in the training set.} %Sequence: left to right, up to down. Notice the successful segmentation in first and last images (negative samples) and second image (a C-shape sample). }
	\label{segress}
\end{figure}

\autoref{segress} shows the segmentation results of a slice stack in the Kaggle training set. Qualitative inspection reveals that the results are satisfactory. The first and last slices exceeded the range of LV, and the model predicted very few pixels belonging to LV. The second slice contained an exit of LV, and the model correctly found this C-shape structure.
For the middle slices, the papillary muscles (dark tissue in LV) were correctly segmented as belonging to LV.

\subsection{Area estimation}
\begin{figure}
	\begin{subfigure}[b]{.48\columnwidth}
		\includegraphics[width=\columnwidth]{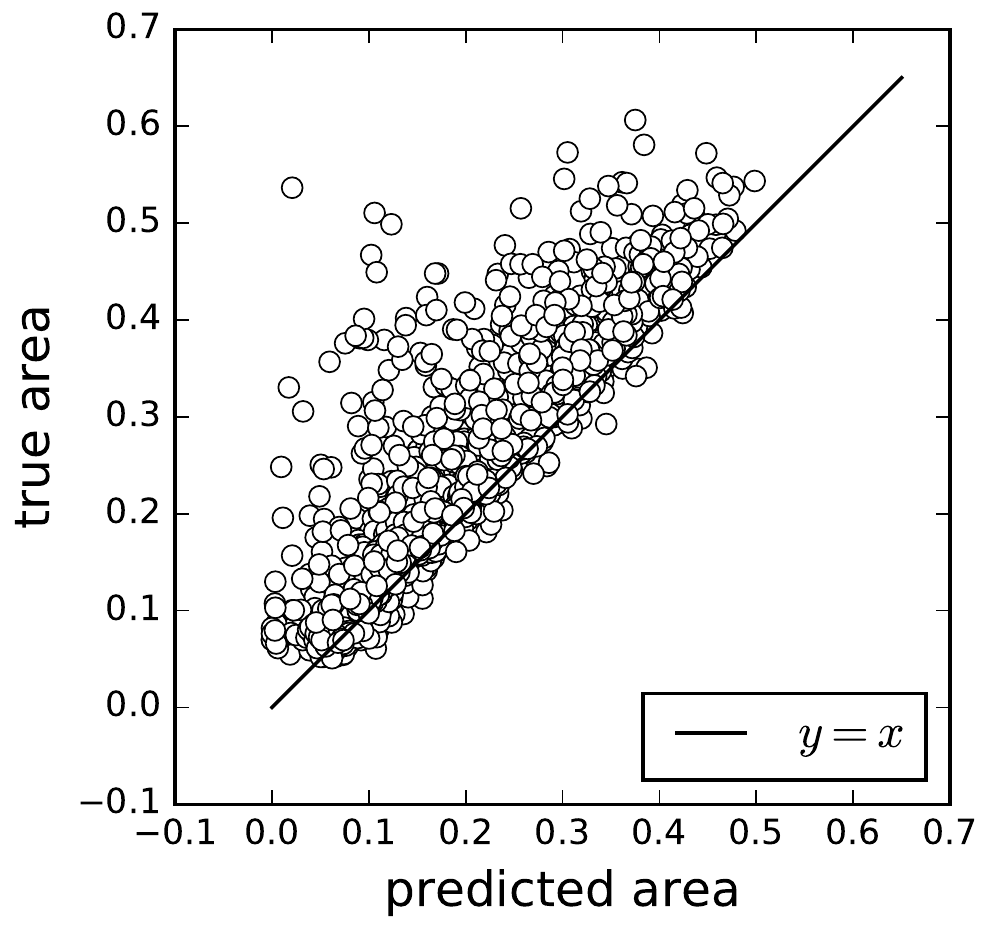}
		\caption{}
		\label{seg1}
	\end{subfigure}
	\begin{subfigure}[b]{.48\columnwidth}
		\includegraphics[width=\columnwidth]{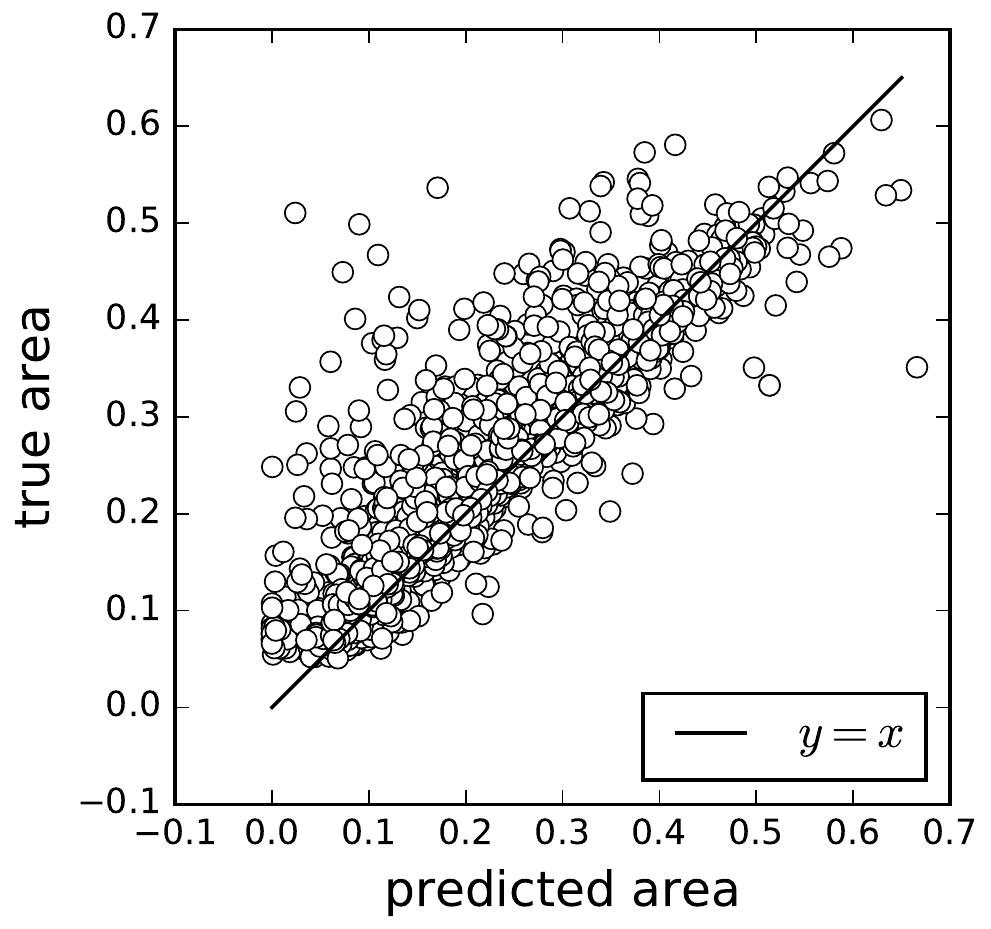}
		\caption{}
		\label{reg1}
	\end{subfigure}
	\vskip\baselineskip
	\begin{subfigure}[b]{.48\columnwidth}
		\includegraphics[width=\columnwidth]{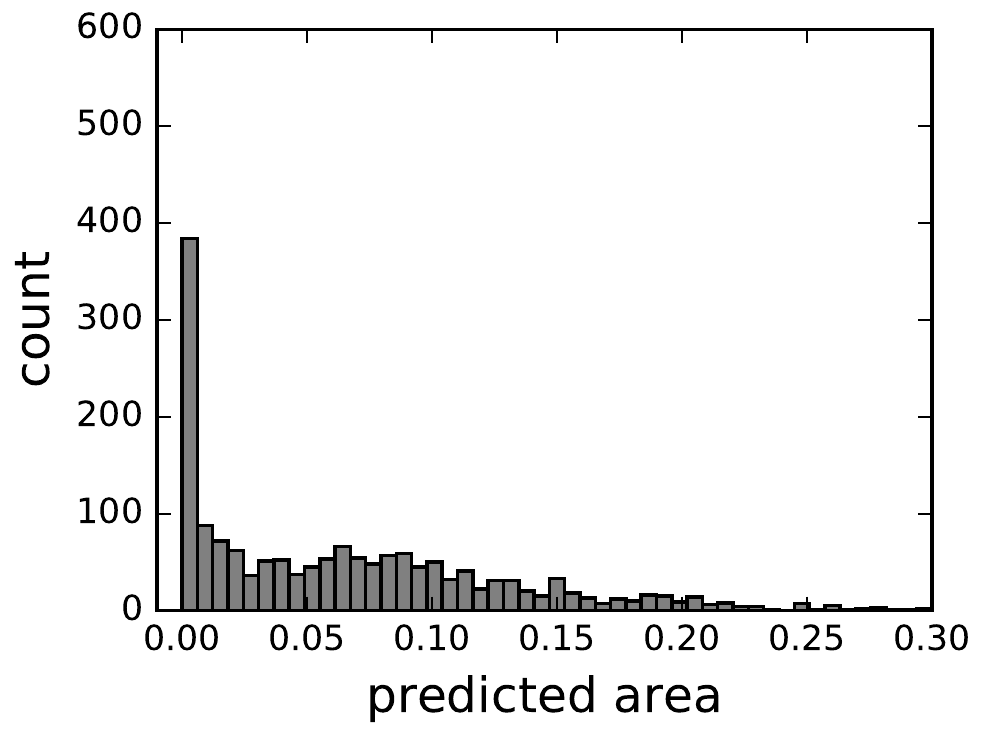}
		\caption{}
		\label{seg2}
	\end{subfigure}
	\begin{subfigure}[b]{.48\columnwidth}
		\includegraphics[width=\columnwidth]{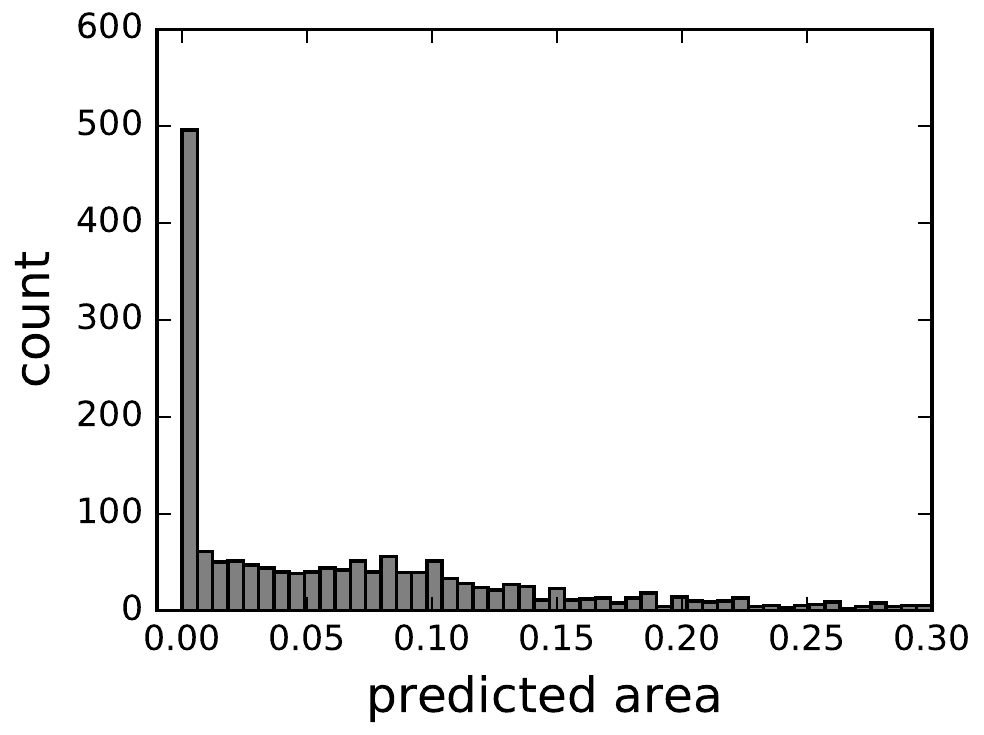}
		\caption{}
		\label{reg2}
	\end{subfigure}

	\caption{Area estimation results of HFCN. (a) and (b) Estimation results on the Sunnybrook dataset using $P$ and $O$, respectively.
		%,x axis denotes the predicted area value, y axis denotes the ground truth, gray line is ideal situation. Most dots area beyond the gray line, which means that most predicted values are smaller than truth.
		%The dots distribute equally across the gray line.
		(c) and (d) The distributions of predicted area of negative samples. using $P$ and $O$, respectively. Note that the first peak in (d) is higher than the first peak in (c).}
	\label{regressResult}
\end{figure}

To evaluate the areas estimation performance of our model, we extracted all available patches from Sunnybrook dataset and augmented the number to 5000 by flipping, scaling, and shifting. We also manually selected 1000 negative samples, but because the authors are not professional doctors, there might be some false negative samples. Then we predicted the areas in these images according to Equations \eqref{singleS} and \eqref{newO} respectively, and compared their predicted values to the ground truth values (ground truths value are calculated from the human-labeled contours). With Equation \eqref{singleS}, the predicted values of positive samples were, on average, smaller than the ground truth (\autoref{seg1}), and the values of some negative samples were greater than 0  (\autoref{seg2}). With Equation \eqref{newO}, both problems were attenuated to certain extent (\autoref{reg1}, \ref{reg2}).

\subsection{Volume estimation}
%As mentioned before, the competition dataset was split into three sets: training, validation and test. The number of these sets were 500, 200 and 440 respectively. Only labels in the training and validation set were available.  As described in Section \ref{section_Data}, for every case there were two frames with volume labels. These frames were used for fine-tuning the model. But on the validation/test sets,

Unlike the procedure in the training stage, where the frames corresponding to diastole and systole state were given, in this stage, we estimated the volumes of all frames in one case and treated the minimum and maximum of these values as ESV and EDV, respectively.

\begin{figure}
	\centering
	\begin{subfigure}{.48\linewidth}
		\includegraphics[width=\linewidth]{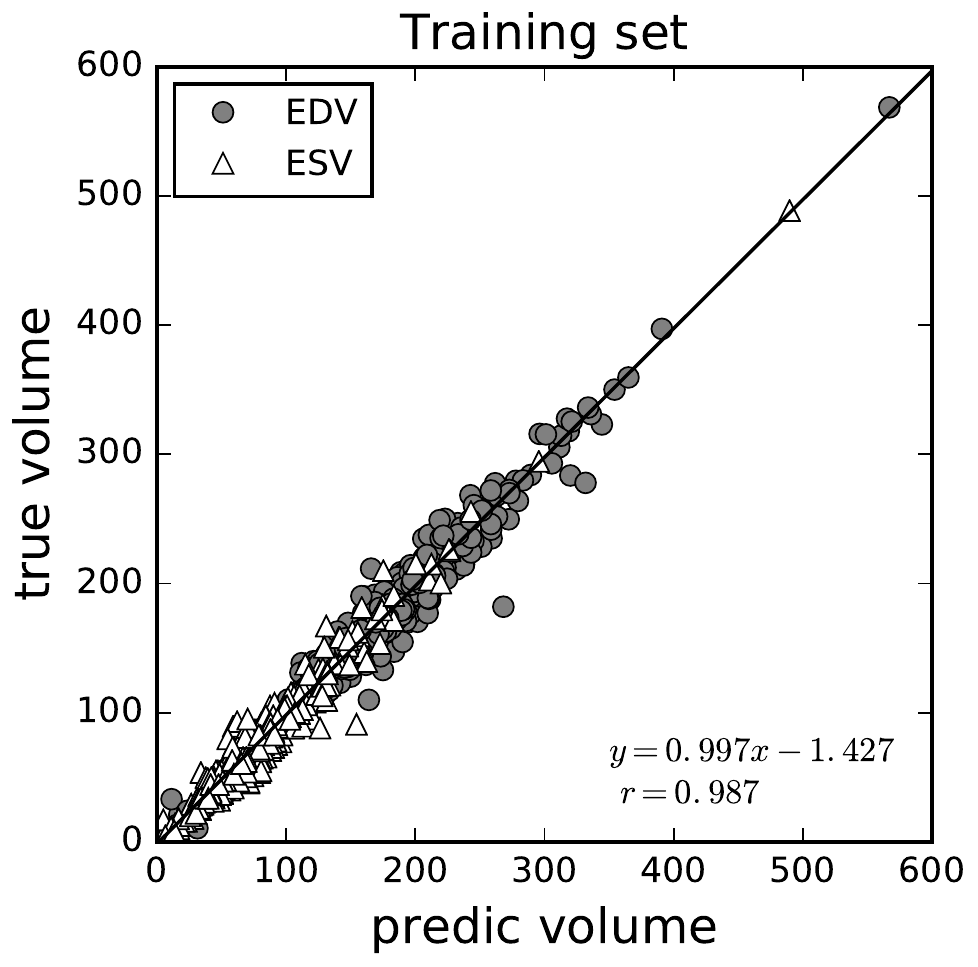}
		\caption{}
		\label{volume_train}
	\end{subfigure}%
	\begin{subfigure}{.45\linewidth}
		\includegraphics[width=\linewidth]{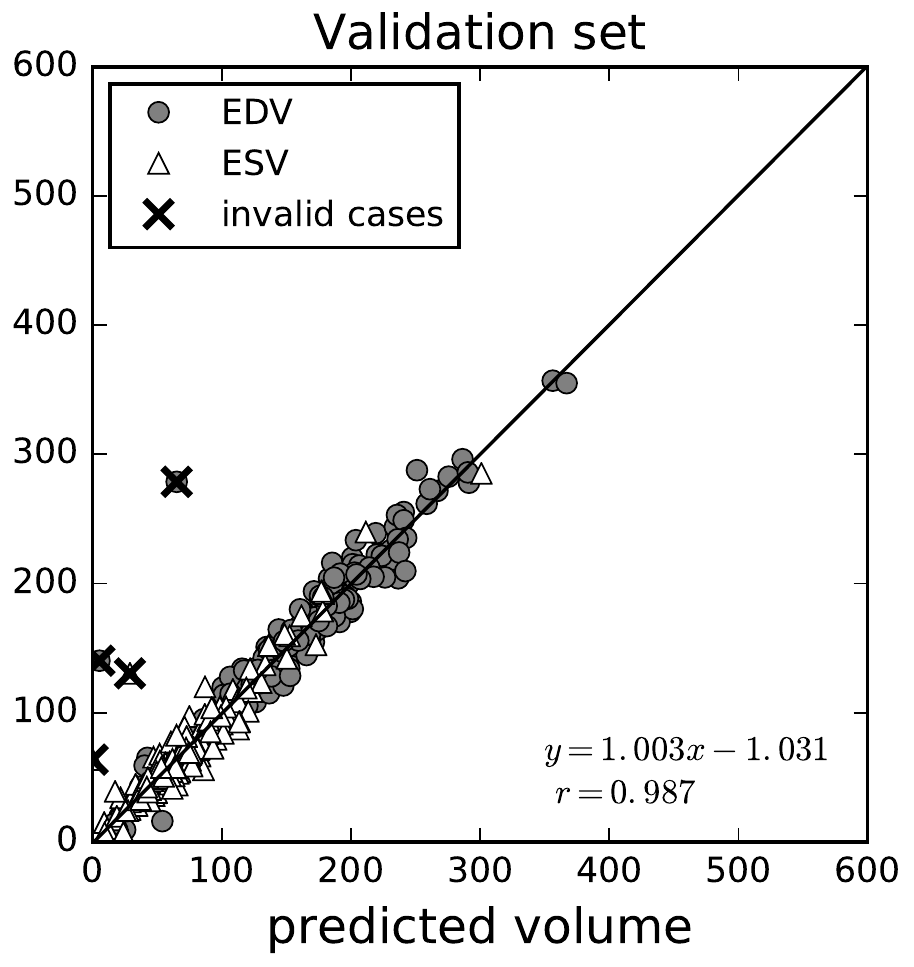}
		\caption{}
		\label{volume_val}
	\end{subfigure}
	
	\begin{subfigure}{.48\linewidth}
		\includegraphics[width=\linewidth]{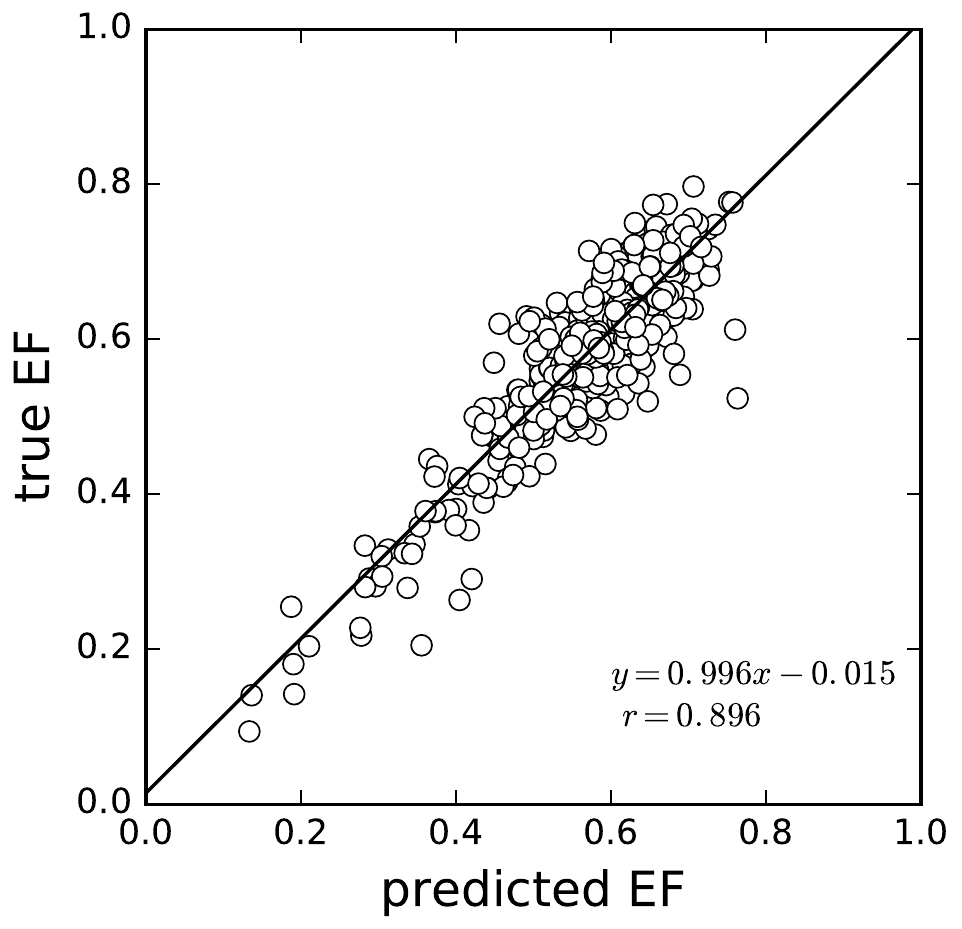}
		\caption{}
		\label{EF_train}
	\end{subfigure}%
	\begin{subfigure}{.45\linewidth}
		\includegraphics[width=\linewidth]{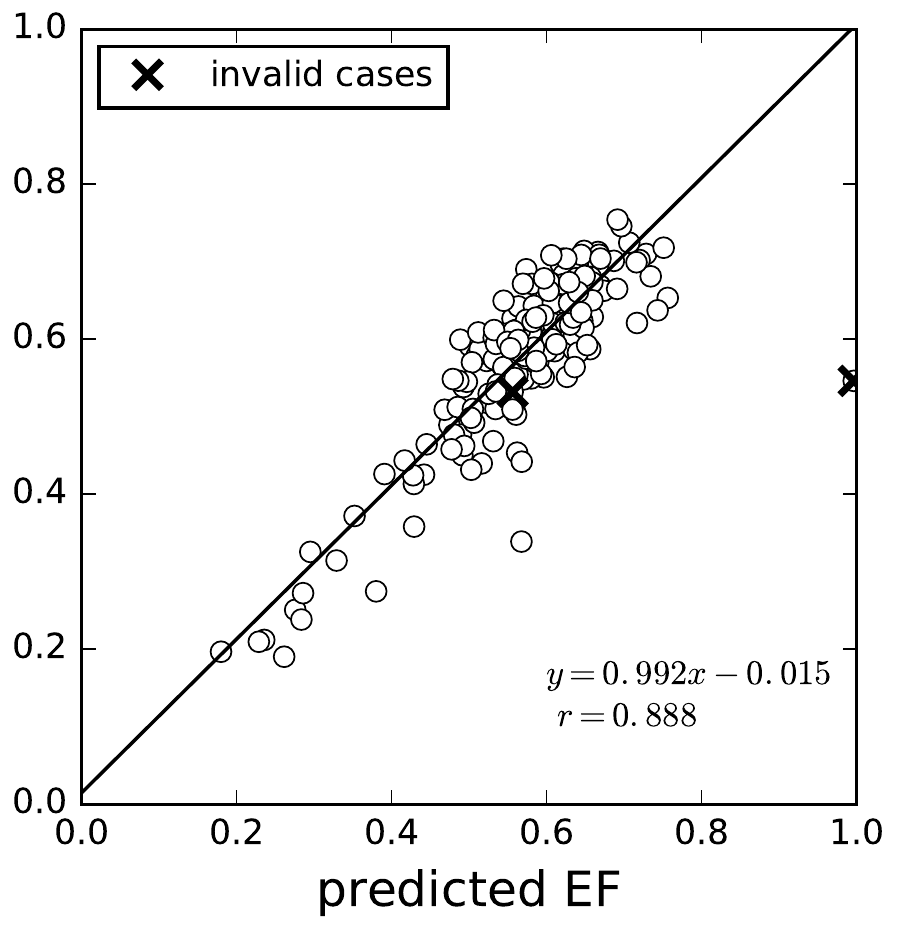}
		\caption{}
		\label{EF_val}
	\end{subfigure}

	\caption{Volume estimation results. (a,b) Estimated EDV and ESV on the training set and validation set, respectively. Each case corresponds to a dot (EDV) and a triangle (ESV).(c,d) Estimated EF on the training set and test set, respectively. Each dot corresponds to one case. Case 595 and 599 are deleted in linear regression. }
	\label{fig:test}
\end{figure}

The results on the training and validation sets are shown in \autoref{volume_train}, \ref{volume_val}. Qualitatively, the regression was good enough, except for some outliers that might be caused by incompleteness of data or incorrect labels. Quantitatively, on the validation set, the average relative error was 13.1\% for ESV and 8.59\% for EDV, and the correlation coefficients between the predicted volume and the ground truth volume were both 0.987. The average $l_1$-norm error of EF was 3.93\% (for regression result, see \autoref{EF_train}, \ref{EF_val}), and the correlation coefficient between the predicted EF and the ground truth was 0.888 (\autoref{EF_val}).

After the competition, the organizer analyzed the results of the leading teams on test set \cite{jonathan_winning_????}. According to the report, in terms of the root mean square error (RMSE) of EF, we achieved the best result in top 4 teams. And in terms of RMSE of EDV and ESV, we also ranked 2nd and 1st, respectively (\autoref{tab_result}).
Unfortunately, the evaluation metric for this competition was not based on any single value estimation but based on the distribution estimation.
% \usepackage{booktabs}
% \usepackage{booktabs}
% \usepackage{booktabs}
% \usepackage{booktabs}
% \usepackage{graphicx}
% \usepackage{booktabs}
% \usepackage{booktabs}
\begin{table}[]
	\centering
	\caption{Evaluation results of the top 4 models in the competition \cite{jonathan_winning_????}}.
	\label{tab_result}
	\begin{tabular}{@{}lcccc@{}}
		\toprule
		& \begin{tabular}[c]{@{}c@{}}mean\\ CRPS\end{tabular}              & \begin{tabular}[c]{@{}c@{}}EDV \\ RMSE\\  (mL)\end{tabular} & \begin{tabular}[c]{@{}c@{}}ESV \\ RMSE \\ (mL)\end{tabular} & \begin{tabular}[c]{@{}c@{}}EF \\ RMSE\\  (\%)\end{tabular} \\ \midrule
		Tencia Woshialex\cite{tencia_summary_????} & \textbf{0.009485} & \textbf{12.02}                                              & 10.19                                                       & 4.88                                                       \\
		kunsthart\cite{ira_diagnosing_????}        & 0.010123          & 13.65                                                       & 10.43                                                       & 6.99                                                       \\
		JuliandeWit\cite{wit_3rd_????}      & 0.010139          & 13.63                                                       & 10.32                                                       & 5.04                                                       \\
		ShowMeTheMoney (ours)  & 0.010666          & 13.20                                                       & \textbf{9.31}                                               & \textbf{4.69}                                              \\ \midrule
	\end{tabular}
\end{table}

\subsection{Evaluation result}

\begin{figure}
	\centering
	\begin{subfigure}{.5\linewidth}
		\centering
		\includegraphics[width=\linewidth]{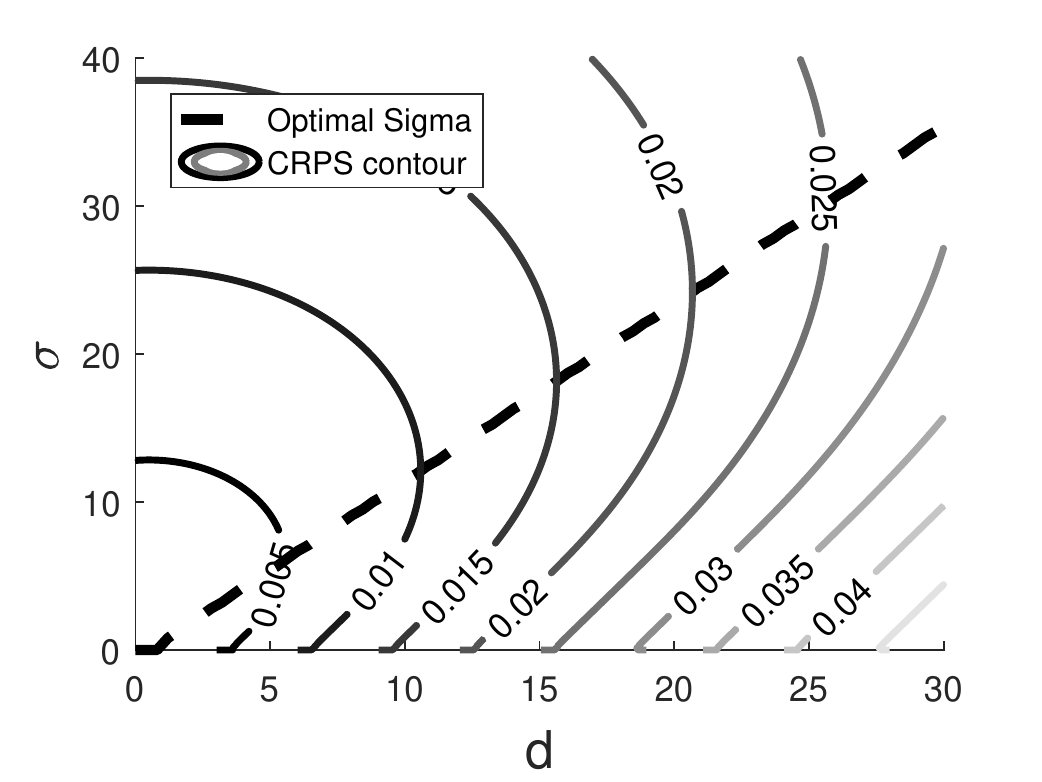}
		\caption{}
		\label{loss}
	\end{subfigure}%
	\begin{subfigure}{.5\linewidth}
		\centering
		\includegraphics[width=\linewidth]{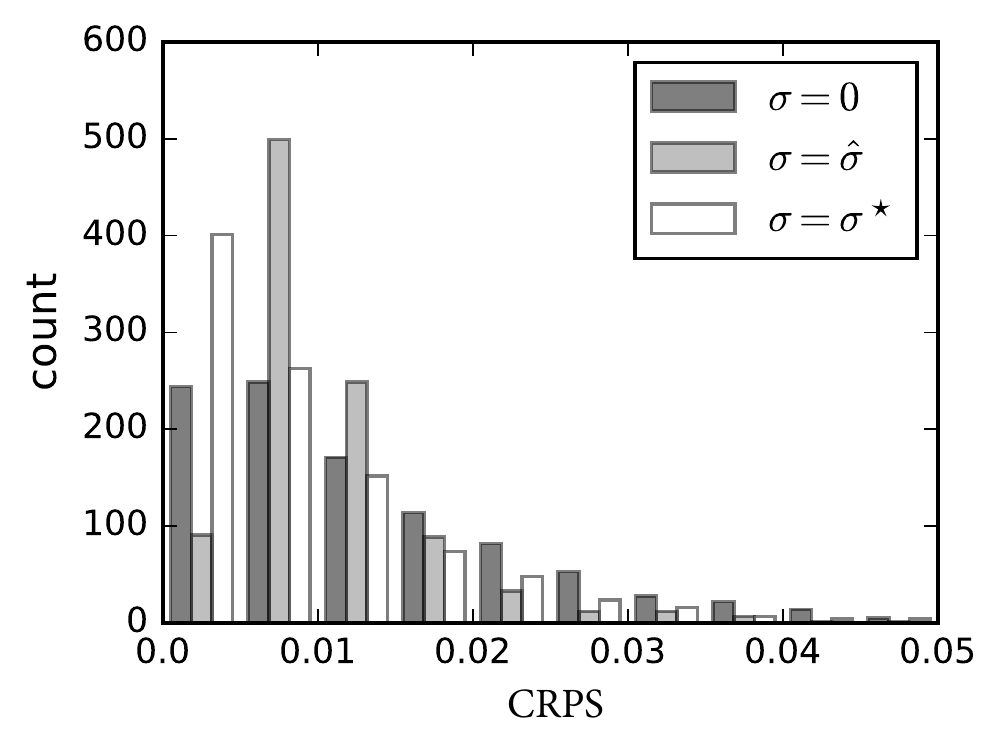}
		\caption{}
		\label{lossresult}
	\end{subfigure}
	
	\caption{Influence of the variance on CRPS. (a) CRPS as a function of $d$ and $\sigma$.  For each $d$, there is a unique optimal $\sigma^\star$ with lowest CRPS. The bold curve plots the function $\sigma^\star(d)$, which is nearly a linear function. (b) Distribution of CRPS on the training set with different $\sigma$.}
\end{figure}

We tested the effect of our variance estimation method (\autoref{lossresulttable}). The distribution of CRPS over the training set is plotted in \autoref{lossresult}. Our method was much better than $\sigma=0$ and was very close to the result of $\sigma=\sigma^\star$. The mean CRPS on the test set was 0.010666, which ranked the 4th in the competition (\autoref{tab_result}). Because we do not know the labels of the test samples, we can not do further analysis.

% \usepackage{booktabs}
\begin{table}[]
	\centering
	\caption{Mean CRPS on the training and validation sets with different variances.}
	\label{lossresulttable}
	\begin{tabular}{@{}llll@{}}

		\toprule
		& Training   & Validation & Test\footnotemark[2]                \\ \midrule
		$\sigma=0$            & 0.01358 & 0.01550    & 0.01400               \\
		$\sigma=\hat{\sigma}$   & 0.00999 & 0.01280    & 0.01067               \\
		$\sigma=\sigma^\star$ & 0.00949 & 0.01112    & \multicolumn{1}{c}{-} \\ \bottomrule
		
	\end{tabular}
\end{table}

\section{Discussion}

%Before this competition, except for handful examples, deep learning is not widely used in biomedical image processing, including this LV segmentation task. But in this competition, deep neural network algorithm shows a big advantage relative to other method: as far as I know, all top 3 team also use deep neural network as their core tool. It means in biomedical image processing, there is still a huge potential of application of deep neural network. After all, the ultimate goal -- automatically diagnosing could always be converted to a classification/detection task in computer vision. In recent years, deep learning enjoyed a fast development due to the vast resources invested, but it hasn't shown its full potential in industry.

%In this work we proposed CFCN, a slightly modified version of FCN. It may not be suitable to general semantic segmentation tasks, but it could be useful in binary segmentation task or those with few training label because of its simplicity.
We used a deep convolutional neural network (DCNN) to segment human MRI images and estimate LV volume. It exhibited good performance in a Kaggle competition. The greatest difficulty in this competition was the lack of informative labels. Our solution was volume fine-tuning and alternate training, which enabled the DCNN to receive supervising signals from both segmentation labels in the Sunnybrook dataset and volume labels in the Kaggle dataset. As far as we know, it is the first 2D segmentation network fine-tuned by utilizing 3D volume labels.

One limitation of our algorithm is that it does not make use of the prior knowledge of the 3D shape of LV, e.g. smoothness of the 3D surface. Neither does it use the smoothness of LV shape in the time domain. As a result, sometimes the model produced segmentation results with two isolated LV regions or a region with a highly irregular shape. This problem can potentially be solved by 3D CNN, or conditional random field as post-processing, which is left for future exploration.

As a powerful tool in computer vision, deep neural network draws more and more attention in biomedical image processing area. As an evidence, in this competition all of the top 4 teams based their algorithms on DCNN. In biomedical image processing, the image features are influenced by lots of parameters: age and gender of patients, viewing angle and position, instrument parameter, and even subject biases of doctors. Traditional methods entail much effort on ruling out those unwanted factors. But DCNN can learn useful features from a large dataset automatically. As long as the training set is large enough to cover all varying factors, DCNN could be robust to them. We believe that DCNN will play more and more significant roles in biomedical image processing.

\appendix
\label{optimal sigma}
\section{Optimal $\sigma$ given prediction variance}

If we parameterize Equation \eqref{CRPS} as $C(\hat{V},V,\sigma)$, according to the definition of $\sigma^\star$:

\begin{displaymath}
\sigma^\star=\arg \min C(\hat{V},V,\sigma).
\end{displaymath}
But in the test stage the exact value of $V$ is unknown. What we have is the estimated value $\hat{V}$ and the estimation variance $\Sigma=\Std(\hat{V})$, based on which we know the distribution of $V$. And we want to have an optimal choice of $\sigma$ so that the expectation of Equation \eqref{CRPS} is minimized:

\begin{displaymath}
\sigma^o =\arg \min \E[C]=\arg \min \int p(V)C(\hat{V},V,\sigma)dV,
\end{displaymath}
where $p(V)$ is the PDF of $N(\hat{V},\Sigma^2)$.

\begin{figure}
	\centering
	\includegraphics[width=\linewidth]{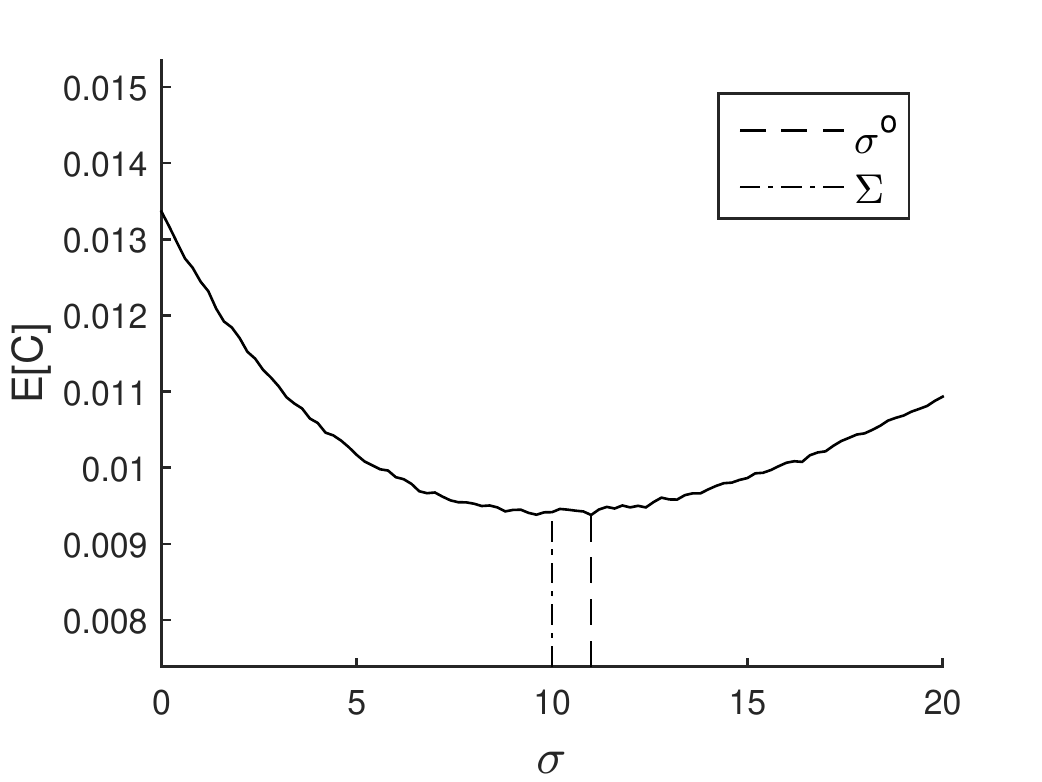}
	\caption{The relationship of $\sigma$ and expectation of CRPS.}
	\label{fig_opti_sigma}
\end{figure}

Since this equation is intractable, we conducted a numerical simulation, i.e. for each $\sigma$, sampled 100,000 $V$ and computed their average CRPS. The result is shown in \autoref{fig_opti_sigma}. It turned out that $\Sigma$ was very close to $\sigma^o$ and their corresponding $E[C]$ were also very close:

\begin{align*}
\sigma^o &\approx \Sigma,\\
\E[C|\sigma=\sigma^o] &\approx \E[C|\sigma=\Sigma].
\end{align*}

In this experiment, $\Sigma$ was set to 10 and $\hat{V}$ was set to 100, which were typical values in this study. We found that this conclusion was robust in a wide range of $\hat{V}$ and $\Sigma$.

%First, $\Std(\hat{V}_m)$ is a straight forward choice of $\sigma$, yet not the optimal one\footnote{The problem is: if $\hat{V}_m \sim N(V_m,\Sigma^2)$, and a fixed $\sigma$ should be chosen for evaluate the \autoref{CRPS} for all $\hat{V}_m$ from this distribution, is $\sigma=\Sigma$ the optimal value? By conducting numerical experiments we found that $\Sigma$ is actually larger than the optimal value.}.


\ifCLASSOPTIONcaptionsoff
  \newpage
\fi



%\bibliographystyle{IEEEtran}
\bibliographystyle{IEEEtranN}
\bibliography{leftventricle}
\end{document}